# Real-Time Roadway Obstacle Detection for Electric Scooters Using Deep Learning and Multi-Sensor Fusion

**Zeyang Zheng, B.E.,[1] Arman Hosseini, M.S.,[2] Dong Chen, Ph.D.,[3] Omid Shoghli, Ph.D.,[4] and Arsalan Heydarian, Ph.D.[5]**

[1] Computer Engineering, University of Virginia, Charlottesville, VA, 22903, USA; e-mail: yuq8cp@virginia.edu
[2] System Engineering, University of Virginia, Charlottesville, VA, 22903, USA; e-mail: ufy5tc@virginia.edu
[3]Agricultural and Biological Engineering, Mississippi State University, Starkville, MS, 39762, USA; e-mail: dchen@abe.msstate.edu
[4]Civil Engineering Technology and Construction Management, University of North Carolina at Charlotte, Charlotte, NC, 28223, USA; e-mail: oshoghli@charlotte.edu
[5]Civil and environmental engineering, Link Lab, University of Virginia, Charlottesville, VA, 22903, USA; e-mail: heydarian@virginia.edu

## ABSTRACT

The increasing adoption of electric scooters (e-scooters) in urban areas has coincided with a rise in traffic accidents and injuries, largely due to their small wheels, lack of suspension, and sensitivity to uneven surfaces. While deep learning-based object detection has been widely used to improve automobile safety, its application for e-scooter obstacle detection remains unexplored. This study introduces a novel ground obstacle detection system for e-scooters, integrating an RGB camera, and a depth camera to enhance real-time road hazard detection. Additionally, the Inertial Measurement Unit (IMU) measures linear vertical acceleration to identify surface vibrations, guiding the selection of six obstacle categories: tree branches, manhole covers, potholes, pine cones, non-directional cracks, and truncated domes. All sensors, including the RGB camera, depth camera, and IMU, are integrated within the Intel RealSense Camera D435i. A deep learning model powered by YOLO detects road hazards and utilizes depth data to estimate obstacle proximity. Evaluated on the seven hours of naturalistic riding dataset, the system achieves a high mean average precision (mAP) of 0.827 and demonstrates excellent real-time performance. This approach provides an effective solution to enhance e-scooter safety through advanced computer vision and data fusion. The dataset is accessible at https://zenodo.org/records/14583718, and the project code is hosted on https://github.com/Zeyang-Zheng/Real-Time-Roadway-Obstacle-Detection-for-Electric-Scooters.

## INTRODUCTION
E-scooters have emerged as a popular and convenient mode of short-distance transportation.

According to a market analysis report ("Electric Scooters Market Size & Share Analysis Report, 2030" n.d.), the global electric scooter market is valued at USD 4.3 billion in 2024, with projections estimating growth to USD 12.4 billion by 2030, at a compound annual growth rate (CAGR) of 18.9%. Despite their growing popularity, the number of e-scooter-related accidents has risen significantly in recent years. In the United States, e-scooter injuries increased dramatically from 8,566 in 2017 to 56,847 in 2022, with hospitalizations from these injuries showing a consistent upward trend (Fernandez et al. 2024).

Several factors contribute to the high incidence of e-scooter-related injuries. The current design of small-wheeled electric scooters often overlooks the complex relationship between wheel size and stability, resulting in improper steering geometries (Paudel et al. 2020). As a consequence, these scooters require higher speeds to achieve self-stability and demand greater rider control to maintain balance at lower speeds. Infrastructure inadequacies exacerbate these challenges, as road obstacles such as potholes and manholes pose significant hazards for e-scooter users. Riders have identified encountering potholes or rough roadways as their most significant concern after being struck by a moving vehicle (Sievert et al. 2023). Additionally, smaller obstacles like pinecones along roadways have been shown to increase riders' visual attention and cognitive load as they navigate to avoid them (Chen et al. 2024b). Moreover, high speeds (>15km/h) greatly increase e-scooter vibrations, which negatively impact comfort and stability (Cano-Moreno et al. 2024). The center of gravity and mass play critical roles in maintaining the e-scooter's stability. Furthermore, a rider's position can shift the center of gravity, indirectly influencing control stability and introducing uncertainty. Compounding these issues, most e-scooters lack shock absorbers that could alleviate severe vibrations. Consequently, electric scooters experience higher vibration frequencies and provide lower comfort levels compared to bicycles (Cafiso et al. 2022).

Deep learning has gained significant attention across various disciplines, particularly in transportation safety-related research. One prominent application is object detection in Advanced Driver Assistance Systems (ADAS), which play a critical role in enhancing the safety of Autonomous Vehicles (AVs) (Gupta et al. 2021). ADAS leverage object detection algorithms to efficiently identify objects such as vehicles, micromobility devices, and road obstacles in real time, while LIDAR sensors measure the distances to nearby objects. Despite the extensive study of object detection in the context of autonomous vehicles, its application to micromobility, particularly e-scooters, remains largely unexplored (Kumar et al. 2025). Recently, YOLO object detectors have gained significant attention due to continuous advancements, making them highly suitable for real-time detection tasks. In Redmon et al. (2016), YOLO has also proven to be a practical approach for computer vision-based object detection in e-scooter applications. Chen et al. (2024a) developed a comprehensive benchmark of 22 selected YOLO object detectors tailored for e-scooter-based applications. Their findings concluded that YOLOv5s (Jocher 2020) offers the best trade-off between efficiency and effectiveness. Furthermore, Li (2020) proposed an algorithm combining depth estimates from a single camera with detection results from the YOLOv3 model to identify obstacles around the scooter. The system issues corresponding prompts based on obstacle distance, achieving a detection accuracy of approximately 70%.

To the best of our knowledge, no prior study has specifically addressed ground obstacle detection for e-scooters. This research utilizes YOLO object detectors to identify roadway obstacles in real time, focusing on six obstacle classes selected based on the significant vibrations they cause to riders, as measured by an IMU sensor. Furthermore, by integrating object detection results with depth data from the depth camera, the system is able to calculate the distance between the e-scooter and detected obstacles.

## METHOD

### Image dataset

The Intel RealSense Camera D435i features an RGB camera, a depth camera, and an IMU, which was mounted on the e-scooter. The mount's angle was adjusted prior to data collection. Figure 1 illustrates the installation of the camera on the Ninebot KickScooter MAX G30LP e-scooter.

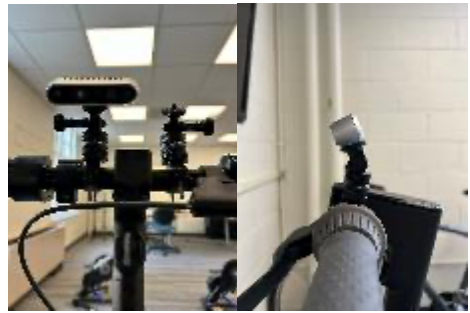

**Figure 1. Camera mounted on the e-scooter**

The selection of roadway obstacle classes draws on findings from prior naturalistic e-scooter riding studies (Chen et al. 2024b) and survey-based research highlighted ground conditions as a significant risk factor for riders (Sievert et al. 2023; Tian et al. 2022). Moreover, the e-scooter's road vibrations are correlated with linear vertical acceleration, which is measured using the accelerometer in the IMU (Trefzger et al. n.d.). Linear vertical acceleration effectively reflects road surface quality. The accelerometer first collects three-axis acceleration data, including components along the $X$, $Y$, and $Z$ axes. Figure 2 illustrates the three axes of the accelerometer within the camera mounted on the e-scooter. Since gravity primarily affects the $Y$ and $Z$ axes, a first-order low-pass filter suppresses high-frequency fluctuations caused by motion, enabling the estimation of gravity components along these axes. The system then subtracts the estimated gravity components from the measured acceleration values to determine the scooter's linear acceleration in the $Y$ and $Z$ directions. To quantify the motion intensity in these directions, the Euclidean norm of the linear acceleration in the $Y$- $Z$ plane is calculated, representing the scooter's vertical linear acceleration:

$$Linear\_vertical\_accel = \sqrt{Linear\_accel\_y^2 + Linear\_accel\_z^2}$$

To mitigate sensor noise, a Kalman filter (Kalman 1960) was applied to smooth the linear vertical acceleration, resulting in a more accurate measurement.

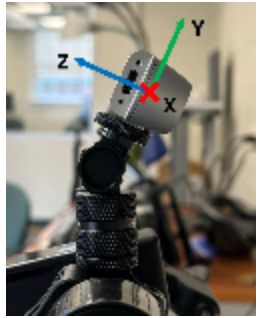

**Figure 2. Three axes of the accelerometer within the camera mounted on the e-scooter**

Experiments were conducted to focus on selecting road hazard classes based on the linear vertical acceleration data from the IMU. Classes that do not cause significant vibrations for e-scooters, such as weeds, were excluded. Figure 3 compares different types of cracks: one generates strong vibrations, while the other produces weak vibrations, as measured by the linear vertical acceleration. Finally, classes selected include manhole covers, non-directional cracks, pinecones, potholes, tree branches, and truncated domes. Figure 4 provides examples of each class.

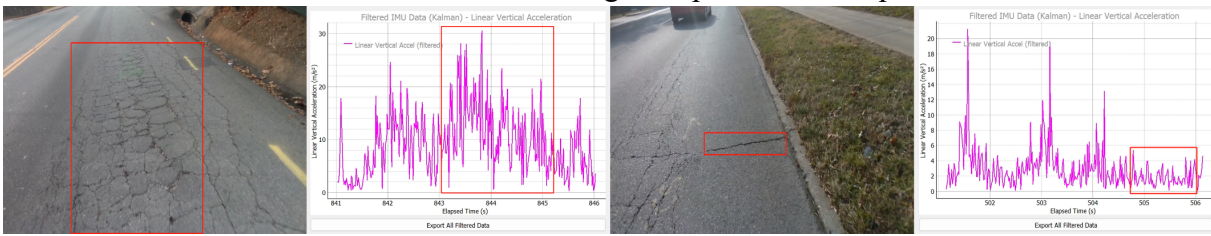

**Figure 3. Linear vertical acceleration: non-directional crack vs vertical crack**

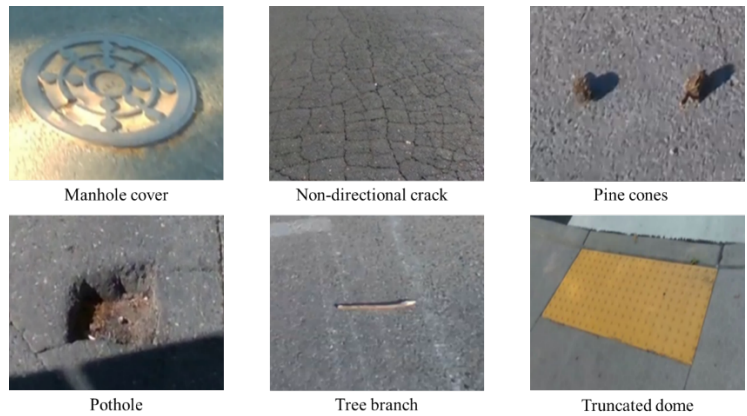

**Figure 4. Examples of six classes of ground obstacles**

Data collection was conducted through natural riding sessions at various daytime hours around the University of Virginia, yielding over seven hours of recorded video footage. The videos were converted into images, and relevant frames were extracted based on classes. Label Studio (Tkachenko et al. 2020) was used to annotate bounding boxes for each class. The resulting dataset comprises 3,427 images categorized into six classes. These road hazards are characterized by highly visible objects rather than less noticeable ones. The dataset includes a total of 8,864 bounding box annotations, categorized as follows: manhole covers (1,050), non-directional cracks (1,310), pinecones (2,866), potholes (1,069), tree branches (1,814), and truncated domes (755).

### Framework

Figure 5 illustrates the complete framework of the ground obstacle detection system for e-scooters. The system utilizes the YOLOv5s model to detect ground obstacles and combines RGB and depth images from the Intel RealSense Camera D435i to estimate obstacle distances in real time.

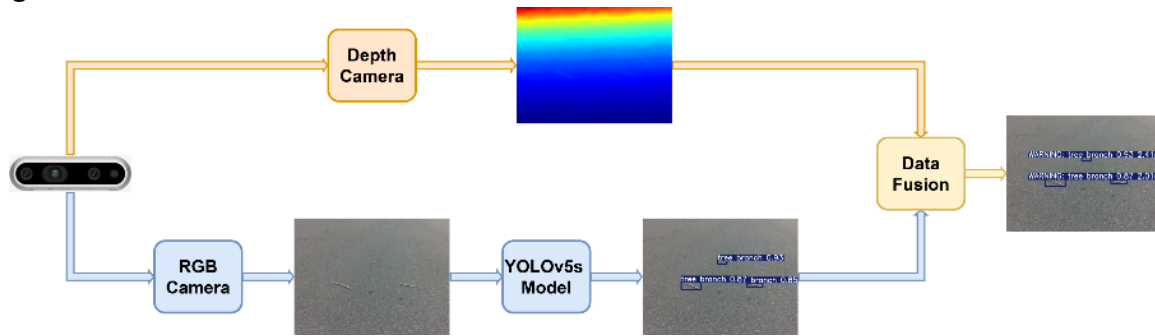

**Figure 5. Framework of ground obstacle detection system for e-scooters**

### YOLOv5s network model

Prior research by (Chen et al. 2024) established that YOLOv5s achieves optimal performance for e-scooter applications. Additionally, comparisons of newer efficient YOLO models, similar in size and complexity to YOLOv5s, are presented in the results and discussion section. This part introduces the architecture of YOLOv5s.

Figure 6 presents the structure of the YOLOv5s network model, which comprises three main components: the Backbone, Neck, and Head. The Backbone forms the foundational component of the network, extracting essential features from input images. The Neck upsamples high-level feature maps and fuses them with low-level feature maps, enhancing semantic richness and capturing small object details. Additionally, low-level feature maps are down-sampled and merged with high-level feature maps, refining detailed information. This bidirectional feature fusion strengthens the network's capability to detect objects across different scales. Finally, the Head processes feature maps of high, medium, and low resolutions from the Neck, enabling the detection of small, medium, and large objects, respectively.

### Fusion of YOLOv5s detection and depth measurement

RGB and depth images are initially aligned using coordinate transformations to ensure that color and depth information for each pixel resides in the same coordinate space. This alignment ensures precise mapping of object depth values to their corresponding color data. After performing target detection on the aligned RGB image using YOLOv5s, the system extracts the bounding box and category information of the detected object. The bounding box provides the coordinates of the object's center point. Due to the presence of noise in the depth data, the system performs random multipoint sampling around the center point to obtain multiple depth values. These values are sorted, and the middle section is selected to calculate the average depth value. This average depth value is then integrated into the detection result, which includes the object category and confidence level. The system demonstrates its ability to notify the rider by using this distance to display real-time warning texts on the detection image. The system outputs a warning text signal on the detection result when the distance falls to 4 meters or less, accounting for the e-scooter's speed and the camera's position on the e-scooter.

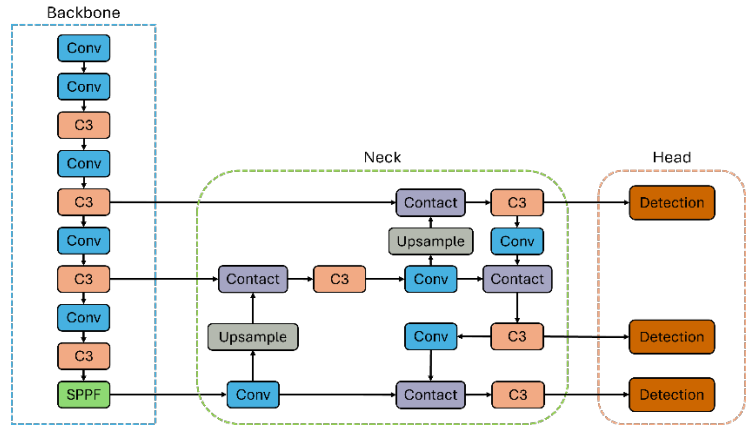

**Figure 6. The structure of YOLOv5s**

**RESULTS AND DISCUSSION**

### Results

To train the YOLOv5s model using the collected data, the input image resolution is set to 1280x1280, batch size is set to 64, and the model is trained for 100 epochs. The dataset was split into training, validation, and test sets in a 7:2:1 ratio. Training took place on a multi-GPU server provided by the University of Virginia Research Computing. The batch size and number of epochs are dynamically adjusted if GPU memory constraints arise. The same trajectory is used to train same-level models from YOLOv6 to YOLOv11(Ao Wang 2024; Jocher et al. 2023; Jocher and Qiu 2024; Li et al. 2022; Wang et al. 2023; Wang and Liao 2024) to compare the performance and efficiency.

Figure 7 presents the YOLOv5s loss function curves for training and validation phases, alongside the performance metrics on the validation dataset across successive training epochs, adhering to the page limit. Precision (P) measures the proportion of true positives among all detected ground obstacles. Recall (R) reflects the model's ability to identify all target classes in the image. The mAP_0.5 (mAP50) represents the average precision computed at an intersection over a union (IoU) threshold of 0.50. The mAP_0.5:0.95 (mAP50-95) calculates the average precision across varying IoU thresholds, from 0.50 to 0.95. These metrics range from 0 to 1, with values closer to 1 indicating superior performance.

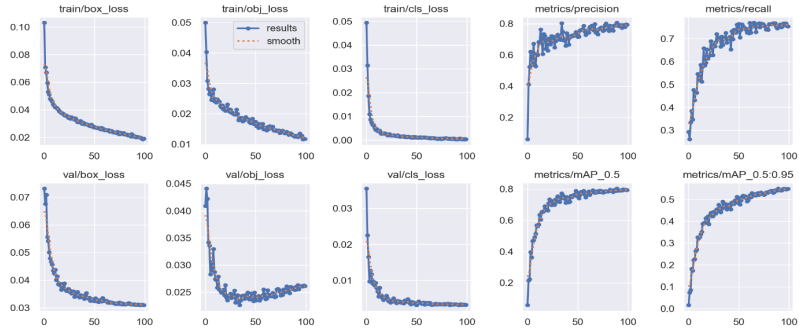

**Figure 7. Training and validation results of YOLOv5s**

Compared to mAP50-95, mAP50 is more practical as it focuses more on detecting target objects (Padilla et al. 2020). Furthermore, mAP50 thoroughly evaluates a model's performance by integrating both P and R. The loss curves demonstrate an overall downward trend for the training and validation datasets, while the performance metrics for the validation dataset show steady improvement throughout the training period.

Table 1 compares the performance metrics, model sizes, and GFLOPs of various YOLO models at the same level on the testing dataset. Model size measures the number of parameters in a model; a smaller model size requires less storage. Giga Floating-point Operations (GFLOPs) serve as a critical metric for evaluating the computational resources required by YOLO models. Lower GFLOPs count signifies reduced computational demands and improved real-time performance. Table 1 shows that the YOLO models demonstrate comparable performance. YOLOv11s achieves the highest precision (P) of 0.838, YOLOv5s records the highest recall (R) of 0.799, and YOLOv9s attains the highest mAP50 of 0.84. Additionally, YOLOv5s has the smallest model size (14 MB) and the lowest computational requirement at 15.8 GFLOPs, which is approximately 74% of the second-lowest model (YOLOv11s) and only 8% of the highest (YOLOv7x). Table 2 highlights the performance of each class detected by the YOLOv5s model on the testing dataset. The truncated dome achieves the highest metrics (P: 0.952, R: 0.983, and mAP50: 0.985). In contrast, the tree branch records the lowest P (0.732), while the non-directional crack exhibits the lowest R (0.556) and mAP50 (0.649).

**Table 1. Performance, model size, and computational cost of YOLO detectors on the testing dataset**

| YOLO models | P | R | mAP50 | Model size (MB) | GFLOPs |
|---|---|---|---|---|---|
| YOLOv5s | 0.834 | **0.799** | 0.827 | **14** | **15.8** |
| YOLOv6s | 0.835 | 0.756 | 0.816 | 38.8 | 45.3 |
| YOLOv7x | 0.809 | 0.778 | 0.807 | 136 | 188.1 |
| YOLOv8s | 0.819 | 0.771 | 0.82 | 21.5 | 28.4 |
| YOLOv9s | 0.834 | 0.788 | **0.84** | 14.6 | 26.7 |
| YOLOv10s | 0.817 | 0.776 | 0.826 | 15.8 | 24.5 |
| YOLOv11s | **0.838** | 0.781 | 0.831 | 18.4 | 21.3 |

**Table 2. Performance by class of YOLOv5s on the testing dataset**

| Class | Images | Instances | P | R | mAP50 |
|---|---|---|---|---|---|
| All | 343 | 831 | 0.834 | 0.799 | 0.827 |
| Manhole cover | 343 | 119 | 0.891 | 0.866 | 0.917 |
| Non-directional crack | 343 | 143 | 0.776 | 0.556 | 0.649 |
| Pine cone | 343 | 220 | 0.837 | 0.884 | 0.894 |
| Pothole | 343 | 124 | 0.816 | 0.694 | 0.725 |
| Tree branch | 343 | 165 | 0.732 | 0.812 | 0.792 |
| Truncated dome | 343 | 60 | 0.952 | 0.983 | 0.985 |

A personal laptop with an Intel i5-13500H processor, NVIDIA GeForce RTX 3050 GPU (6GB), 16GB of RAM, and Windows 11 serves as the test device. Inference size is configured to 640×480 to better focus on the road ahead of e-scooters. The system achieves an average frame rate exceeding 100 FPS. Figure 8 shows the system's detection results for ground obstacles positioned beyond and within 4 meters of the electric scooter.

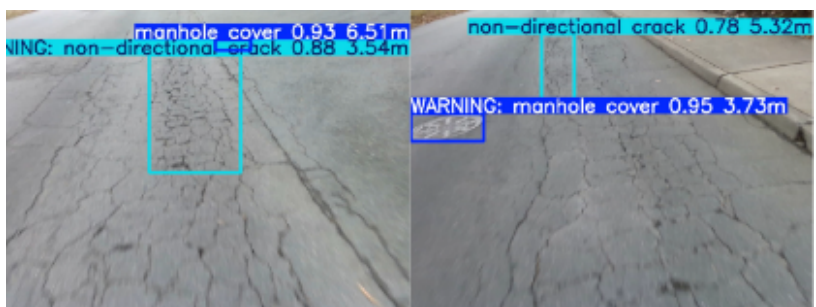

**Figure 8. Examples of the system detection results**

**Discussion**

Compared to other same-level YOLO models, YOLOv5s achieves superior computational and storage efficiency while maintaining similar performance. These results show that YOLOv5s is better suited for e-scooters, which are devices with limited computing resources. In the future, deploying the trained YOLOv5s model on an embedded device in e-scooters will be both feasible and efficient.

A comprehensive analysis of YOLOv5s metrics for each category on the test dataset reveals notable performance variations. The manhole covers and truncated dome categories achieve excellent mAP50 scores of 0.917 and 0.985, respectively, likely due to their large sizes and distinctive features, which make them easily recognizable. The pinecone and tree branch categories contain more instances than others, as the dataset emphasizes collecting smaller target objects. However, the tree branch's mAP50 (0.792) is significantly lower than that of the pinecone (0.894). The pothole category records a mAP50 of 0.725, while the non-directional crack has the lowest mAP50 of 0.649. For the tree branch, pothole, and non-directional crack categories, the low mAP50 values are likely due to varying shadows and light reflections, which make them harder to recognize (Diwan et al. 2023). Improving detection performance requires diverse data collection across various categories under different conditions, such as nighttime and rainy weather.

Adopting advanced or alternative detection models and refining training methodologies may help address these challenges more effectively. Additionally, expanding datasets to include additional categories, such as stones, could enhance safety and usability for e-scooter riders.

The system demonstrates a robust ability to measure distances to detected objects and explores its potential to warn e-scooter riders by displaying real-time warning messages.

## CONCLUSION

The increasing adoption of e-scooters in urban environments has led to a rise in e-scooter-related crashes and injuries. Due to design limitations, e-scooters are particularly susceptible to vibrations caused by road hazards, making rider safety on uneven roadways a pressing concern. To address this issue, this paper proposes a real-time ground obstacle detection system for e-scooters. The system integrates an RGB camera and a depth camera to accurately and efficiently detect six types of road obstacles, achieving an overall high mAP50 of 0.827 while maintaining computational cost-efficiency. The IMU measures vibrations through linear vertical acceleration, which informs the selection of these six obstacle types. These obstacles include manhole covers, non-directional cracks, and potholes, identified using YOLO object detection models. By fusing RGB data and depth data, the system facilitates precise distance estimation. However, further inclusion of additional road obstacle classes harmful to e-scooters is necessary, and the system's performance still has room for improvement. Future work could focus on deploying the system on embedded devices with real-time notification systems, such as voice alerts or dashboard displays. This research presents an effective real-time solution for detecting ground obstacles encountered by e-scooters, paving the way for advancements in smart e-scooter safety and contributing to the broader development of smart mobility.